\documentclass{article}
\usepackage{graphicx}
\usepackage{authblk}
\usepackage{amsmath}
\usepackage{multirow}
\usepackage{tablefootnote}
\usepackage{array}
\usepackage{setspace}
\usepackage[table,xcdraw]{xcolor}
\usepackage{lineno}
\usepackage{hyperref}

\usepackage[sorting=none]{biblatex}

 \title{Deep Neural Networks with 3D Point Clouds for Empirical Friction Measurements in Hydrodynamic Flood Models}
\author[1,2,*]{Francisco Haces-Garcia}
\author[2]{Vasileios Kotzamanis}
\author[1,2]{Craig L. Glennie}
\author[2]{Hanadi S. Rifai}
\date{March 2024}

\affil[1]{National Center for Airborne Laser Mapping, 5000 Gulf Freeway, Houston, TX 77204-5059}
\affil[2]{Department of Civil and Environmental Engineering, University of Houston, Engineering Building 1, Room N107, 4226 Martin Luther King Boulevard}
\affil[*]{Corresponding Author: fhacesgarcia@uh.edu}

\begin{document}

\maketitle

\section{Abstract}
Friction is one of the cruxes of hydrodynamic modeling; flood conditions are highly sensitive to the Friction Factors (FFs) used to calculate momentum losses. However, empirical FFs are challenging to measure because they require laboratory experiments. Flood models often rely on surrogate observations (such as land use) to estimate FFs, introducing uncertainty. This research presents a laboratory-trained Deep Neural Network (DNN), trained using flume experiments with data augmentation techniques, to measure \textit{Manning's n} based on Point Cloud data. The DNN was deployed on real-world lidar Point Clouds to directly measure \textit{Manning's n} under regulatory and extreme storm events, showing improved prediction capabilities in both 1D and 2D hydrodynamic models. For 1D models, the lidar values decreased differences with regulatory models for in-channel water depth when compared to land cover values. For 1D/2D coupled models, the lidar values produced better agreement with flood extents measured from airborne imagery, while better matching flood insurance claim data for Hurricane Harvey. In both 1D and 1D/2D coupled models, lidar resulted in better agreement with validation gauges. For these reasons, the lidar measurements of \textit{Manning's n} were found to improve both regulatory models and forecasts for extreme storm events, while simultaneously providing a pathway to standardize the measurement of FFs. Changing FFs significantly affected fluvial and pluvial flood models, while surge flooding was generally unaffected. Downstream flow conditions were found to change the importance of FFs to fluvial models, advancing the literature of friction in flood models. This research introduces a reliable, repeatable, and readily-accessible avenue to measure high-resolution FFs based on 3D point clouds, improving flood prediction, and removing uncertainty from hydrodynamic modeling. 
\section{Highlights} 
\begin{itemize}
    \item Introduces method to measure Manning’s n from 3D point clouds, improving flood models
    \item Laboratory-trained Deep Neural Network achieves better measurements than land cover
    \item Provides a new, low-cost, and repeatable alternative to measure roughness in models
    \item  Reveals impacts of friction factors under fluvial, pluvial, and surge flooding
    \item Improves flood models of Hurricane Harvey in Houston, per imagery and gauge data
\end{itemize}
\section{Introduction}
Hydrodynamic modeling is a crucial tool in the prediction and mitigation of the hazardous impacts of flood events. The importance of friction estimation for both 1D and 2D hydrodynamic models has been shown in the literature, with these parameters causing significant changes to predicted flood conditions (e.g., \cite{AbuAly2013}, \cite{Dorn2014}, \cite{Ozdemir2013} ). However, quantifying these distributed momentum losses is challenging, particularly because the underlying phenomena are dependent on both scale and flow conditions \cite{Lane2005}. Because of the difficulties in friction quantification, the standard practice is to use friction factors (FFs). FFs are empirically-derived values which can be applied to estimate momentum losses by considering flow conditions; some examples include \textit{Manning's n}, \textit{Strikler's K}, and \textit{Chezy's C}. The reader is referred to \cite{Smith2007} for a comprehensive review of the equations used to estimate flow friction, and their corresponding FFs. Because of their relative simplicity and longstanding use, FFs are commonly implemented in hydrodynamic flood models. \par
Traditional approaches for deriving FFs (such as practical experience and reference manuals) are still in use, however, they are resource-intensive, highly subjective and uncertain. Moreover, they are susceptible to bias due to varying measurement criteria across locations and time frames. An alternative has been the use of remote sensing, which, at the time of writing, remains one of the most common techniques to estimate FFs in hydrodynamic modeling \cite{medeiros2012, Dorn2014}. Standard practice is to obtain surrogate surface information derived from image data, and convert it to standardized FF values. In some hydrodynamic models, in fact, it is possible to directly input Land Use, Soil Classification, and other derivatives from remote sensing to select FFs \cite{Brunner2010, TUFLOW}. However, there are several well-documented limitations to such approaches, including coarse resolution, intraclass variability \cite{AbuAly2013, medeiros2012, Medeiros2015}, and the lack of in-channel information \cite{Gibson2021}. Therefore, novel data analysis techniques have been sought in the literature to make better use of the available data.\par
% \subsection{Machine Learning for F riction Factors}
Machine Learning has often been used in the estimation of FFs. Approaches are varied, with earlier works using flow parameters to estimate in-channel FFs (e.g. \cite{Lopez2002}). More recent works have developed decision support systems powered by machine learning \cite{Niazkar2019}. Deep Neural Networks (DNNs) have been applied to update FFs dynamically \cite{Fu2016}, and estimate them based on flow conditions \cite{Yarahmadi2023}. DNNs have also been used on remote sensing imagery to directly estimate FFs \cite{mehedi2022}. Unfortunately, these approaches generally rely on imagery and in-situ flow information, which only partially capture the true surface characteristics that cause friction.\par
 % \subsection{Lidar for friction factors}
Point Cloud (PC) data has long been regarded as a potential solution to FF estimation \cite{Lane2005, Smith2007, Forzieri2012}, since they are 3D representations of surface topography and roughness, which are directly related to bottom friction \cite{Lane2005}. PCs are widely available from high-resolution lidar or photogrammetric surveys; coupled with their potential usefulness to friction estimation, this constitutes them as the most underutilized data source for FF estimation. Applications of PCs for FF estimation in the literature have been limited. Notably, \cite{Straatsma2008} used PCs and multispectral data to measure vegetation density and land cover, which were then used to estimate FFs as inputs in a hydrodynamic model. \cite{Medeiros2015} developed a Random Forest Regressor based on statistics from the lidar PC and showed a 53\% improvement over Land Use-based values to predict Manning's n.\par
Despite these encouraging results using PCs to measure FFs, significant gaps in the literature remain. No research to date has implemented raw 3D PCs to directly measure FFs for a hydrodynamic model. Moreover, even though FFs are empirical values, no research has harnessed DNNs to translate laboratory measurements of FF into real hydrodynamic models. Furthermore, the implementation of PC-based FFs in hydrodynamic models has been limited, as such, the usefulness of such approaches for various types of flooding remains largely unstudied. To mitigate these methodological, practical, and knowledge gaps, this research develops a laboratory-trained DNN to predict FFs, and applies it in steady and unsteady environments to model regulatory, fluvial, pluvial, and surge flooding.\par
\section{Methods}
To develop, train, and apply a Deep Neural Network for FF estimation in hydrodynamic models there were several practical considerations. Without loss of generality to other FFs, \textit{Manning's n} \cite{manning1890} was selected because of its widespread use in hydrodynamic flood models. To foster repeatability and expand applicability of this study, open source and freely-available software packages were preferred to commercial options. As such, all hydrodynamic flood modeling was conducted in HEC-RAS 6.2, a state-of-the-art, freely-available modeling platform developed and distributed by the US Army Corps of Engineers \cite{Brunner2010}. HEC-RAS is commonly used in the literature, and widely implemented for regulatory flood mapping in the United States.  The remainder of the methods section is structured as follows: first, the laboratory measurement and remote sensing of \textit{Manning's n} is described, then, the development of a DNN based on those results is detailed, and finally, the deployment and assessment of the DNN in HEC-RAS is presented.
\subsection{Experimental Measurement of Friction Factors}
\begin{figure}[!htb]
	\centering
	\includegraphics[width=1\textwidth]{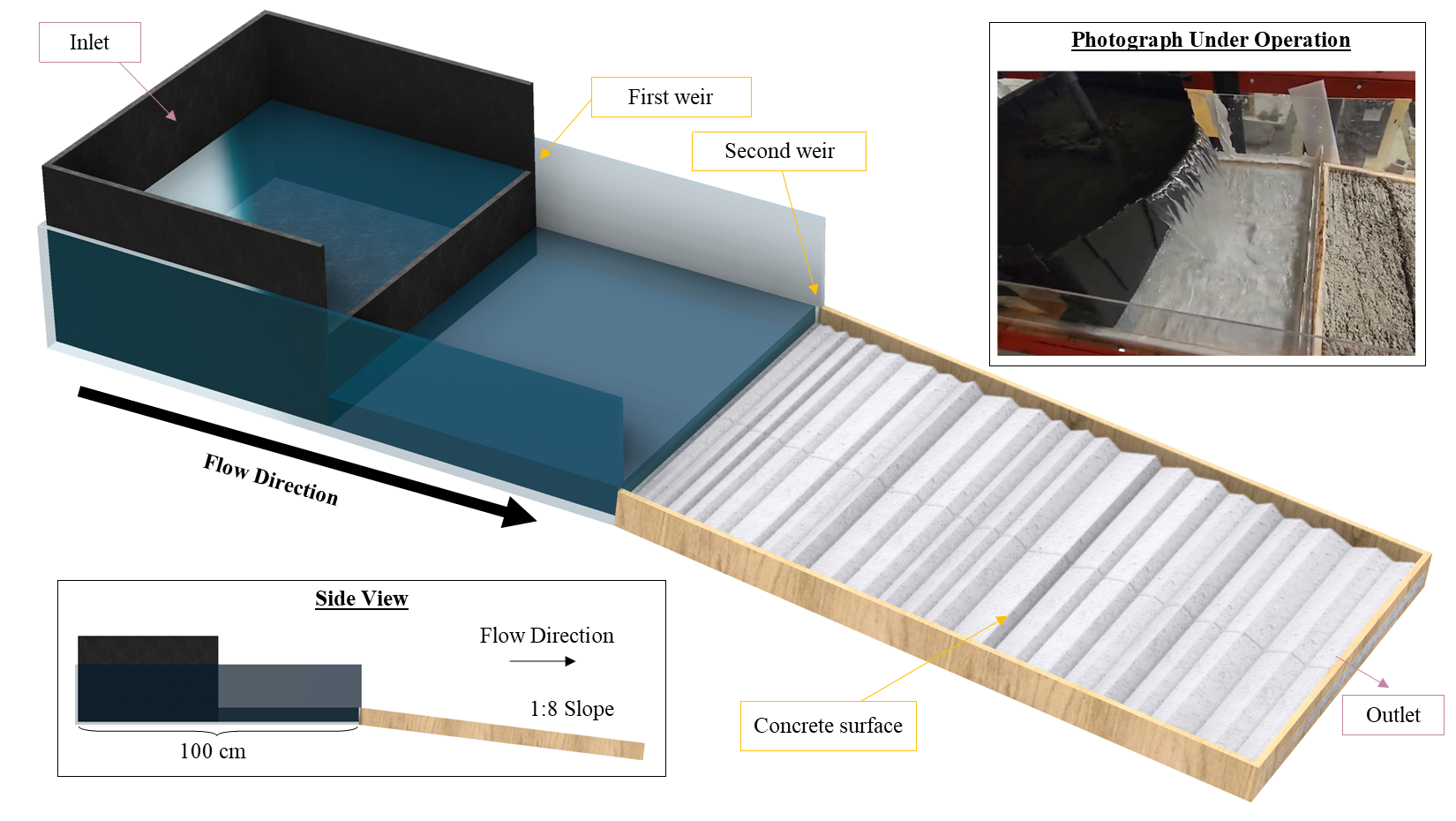}
	\caption{Isometric and side view of weir system used for uniform flow distribution. A dual weir was implemented to ensure near-zero velocity at hydraulic jump, and to remove flow turbulence from the final weir.}
	\label{weir}
\end{figure}

A 10-m non-recirculating flume located at the University of Houston South Park Annex was utilized in this study to measure \textit{Manning's n}. The flume had a wave generator and a foam-based 1:8 slope to model wave run-up. To adapt it for this study, a submersible pump (Lowara DOC3/A) was used in the drainage sump to provide a constant flow rate at the top of the slope, which returned to the partially-filled flume downstream. A suppressed rectangular weir system was designed and fabricated to uniformly distribute flow, whose flow-rate was manually adjusted by a series of valves at the inflow. This setup is shown in Figure \ref{weir}. \par

Three concrete slabs were cast with various surface patterns to model a variety of friction conditions. They are referred to as "Rough", "Medium", and "Smooth" which indicates their relative surface roughnesses. The slabs had measurement ridges to allow several conductivity-based twin wire free surface elevation probes (HRIA-1013) to sit flush with the surface. The probes recorded water depth at a frequenzy of 500 Hz. These probes return an analog voltage which is directly proportional to the immersed depth \cite{WaveProbe}. They were calibrated using the included software, HR DAQ (HRSW-1069) \cite{HRDAQ}, by submersing them at known depths and recording the resulting voltages.\par
A Nixon Streamflo 403 Channel Velocity meter (SN: 3305) was used to measure water velocity, along with a handheld Nixon 440 Digital Measurement Indicator. The digital display (as well as a separate Nixon digital display for the sump pump flowrate) were recorded using a commercial handheld camera, and later transcribed. The Nixon channel velocimeter was unable to achieve a stable measurement in high-velocity flow conditions (over 20 cm/s) due the small inundation depth. Thus, Particle Tracking Velocimetry (PTV) was additionally implemented as an alternative velocity measurement. The PTV recordings were created by the simultaneous use of three GoPro Hero10 Blacks, recording at 240 frames-per-second, and the introduction of buouyant particles made from styrofoam.\par
\begin{figure}[!htb]
	\centering
	\includegraphics[width=1\textwidth]{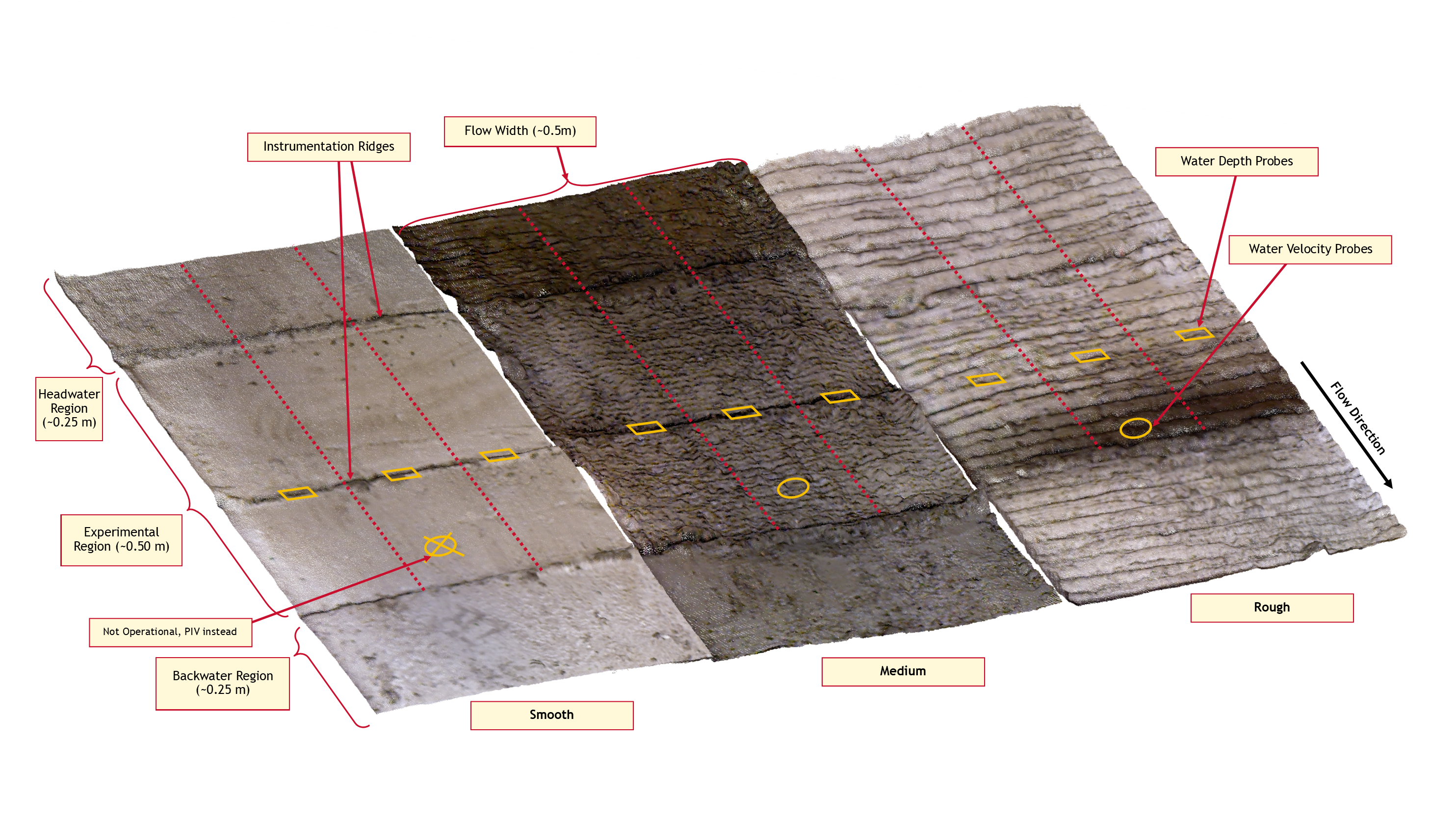}
\caption{Experimental setup to measure \textit{Manning's n}. As shown, the smooth surface had low water depths, resulting in difficulties measuring water velocity with the Channel Velocity meter. Therefore, PTV was used. Figure modified from \cite{AGUPres}}
	\label{experiment}
\end{figure}

To obtain \textit{Manning's n} measurements from the concrete slabs, they were placed downstream of the weir on the slope, and measurements for water depth and velocity were collected for 180s at several input flow rates after flow had stabilized. Under test conditions, a maximum flowrate over the weir of 60 liters-per-minute could be achieved; the lowest flowrate was established by the limits-of-detection for both the depth and velocity probes, which varied across the concrete slabs, but was consistent for the measurement regions within each slab. To account for the heterogeneity of the surfaces due to manual casting, three separate lateral zones were created within each concrete slab, for which water depth was measured. These are named as "Left", "Center", and "Right", indicating their relative locations according to the downstream flow direction. This setup is shown in Figure \ref{experiment}. One velocity measurement was used for all three zones, both due to the rectangular weir controlling the uniformity of flow, and the practicability of monitoring the channel velocimeter to ensure the proper capture of velocity. A total of nine zones (three for each of the three slabs) were measured with at least four different flow rates, providing a total of nine \textit{Manning's n} measurements based on at least four flow rates each. 

\subsubsection{Collection of Remote Sensing Data}
Two remote sensing methods were used to obtain 3D PCs from the concrete castings. The first was a PrimeSense Carmine 1.09 (SN: 201411556), which is a handheld dot-projection lidar scanner. This scanner was operated by an NVIDIA Shield (MN: P1761W) mounted on a gimbal. To scan the concrete slabs, sequential passes were performed in orthogonal directions, followed by additional oblique views to maximize PC coverage. Following this, with the scan still operational, coverage maps provided by PrimeSense's software were analyzed to ensure an approximately homogeneous distribution of point returns. After scanning, the PCs were post-processed and exported using PrimeSense's software on the NVIDIA Shield. Final point cloud densities for the handheld scanner varied from $1.9e6$ to $4.3e6$ $points/m^2$.\par
The second remote sensing technique was using a flume-mounted bed profiling system (BPS) by HR Wallingford (HRBP-1070). This system consisted of a traverser able to move in 3 dimensions across the flume bed surface, and a laser-based profiler in a waterproof casing to measure distance. The BPS is driven by a command workstation, which can set scanning patterns and record measurements. Conventional operation calls for an interlink between the traverser and the laser profiler such that the latter is kept at a constant distance from the bed surface, and the bed profile can be extracted from the traverser position. Due to the rapidly-changing nature of some of the concrete slabs, however, this proved a challenge, which was overcome by developing a calibration curve between the laser sensor's voltage and it's distance to the target. Such measurements were converted to a point cloud by considering the traverser position at sensing time, and the laser voltage through the calibration curve. Final point cloud densities for the traverser were $4.4e5$ $points/m^2$.\par
\subsubsection{Calculation of Manning's n}
Using the collected measurements from each experiment, \textit{Manning's n} was calculated for each region using equation \ref{eq:manning}, where $S$ is the bottom slope, $V$ is the flow velocity, and $R$ is the hydraulic radius calculated by equation \ref{eq:hydraulic} for a rectangular cross section of height $h$ and width $w$. Note that equation \ref{eq:manning} is equivalent to the SI version of Manning's Equation. \par
\begin{equation} \label{eq:manning}
n=\frac{R^{2/3}*S^{1/2}}{V}
\end{equation}
\begin{equation} \label{eq:hydraulic}
R= \frac{h * w}{2 * h + w}
\end{equation}
\subsection{Deep Neural Network Engineering} 
Once the measurements of Manning's n had been obtained, a DNN architecture capable of predicting them based on variable-length PCs was required. PointNet \cite{PointNet} was selected for its proven efficacy in semantic segmentation and classification tasks while operating on raw 3D PCs. For the purposes of this study, it was adapted to be used in a regression task.\par
\begin{figure}[!htb]
	\centering
	\includegraphics[width=1\textwidth]{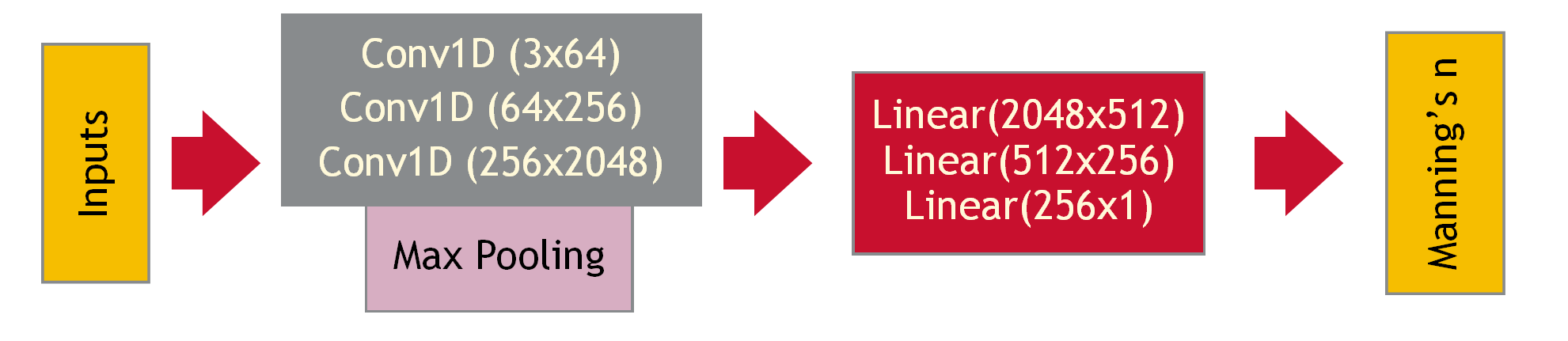}
	\caption{Implementation of PointNet architecture. Conv1D stands for a 1D convolution with a Batch Normalization layer, and linear for linear layers with rectified linear unit nonlinearities. Figure adapted from \cite{AGUPres}.}
	\label{pointnet}
\end{figure}

% \subsubsection{Data Augmentation and Preprocessing}
Since the nine Manning's n measurements collected were insufficient to train a DNN, data augmentation techniques were required. The first was random PC subsampling, in which between three and 100 points from the PCs were randomly taken as samples. After this, all PCs were normalized to have a zero-origin. Finally, data blending was implemented by randomly merging two subsamples of the normalized PCs, and calculating a corresponding weighted \textit{Manning's n} through equation \ref{eq:merge}. 
\begin{equation} \label{eq:merge}
n_{new} = \frac{n_1 * size(PC_1) + n_2 * size(PC_2)}{size(PC_1) + size(PC_2)}
\end{equation}
% \subsubsection{Architecture Tuning and Training}
To adapt the PointNet architecture to a regression task, the decoder connections were replaced with a series of feedforward layers that ultimately provide a single prediction for each processing batch. Other variable-length decoding strategies were explored, but showed limited improvement. The size of the kernels and hidden sizes in the encoding and decoding connections were also changed. The transformation networks were removed to decrease the number of parameters in the model (and thus stabilize training). Instead, the PCs were normalized prior to inference and training, in accordance to the data augmentation techniques previously discussed. The adaptations made to the PointNet are shown in Figure \ref{pointnet} and the underlying code is publicly available at \href{https://github.com/f-haces/LidarManning}{https://github.com/f-haces/LidarManning}. \par
DNN training was performed using the Adam optimizer \cite{kingma2017adam}, with an L1 Loss function and a scheduled learning rate decay. To improve gradient calculations, \textit{Manning's n} values were upscaled by a factor of 1e5. Hyperparameter tuning was performed, and ultimately, the selected DNN setup had 34.3 million parameters, and was trained on a workstation-grade 8GB NVIDIA M4000 until convergence. Outputs were postprocessed by thresholding between 0.025 and 0.25, which was taken as the effective range of the measurements in the underlying experimental data. \par
\subsection{Hydrodynamic Modeling}
\begin{figure}[!htb]
	\centering
	\includegraphics[width=1\textwidth]{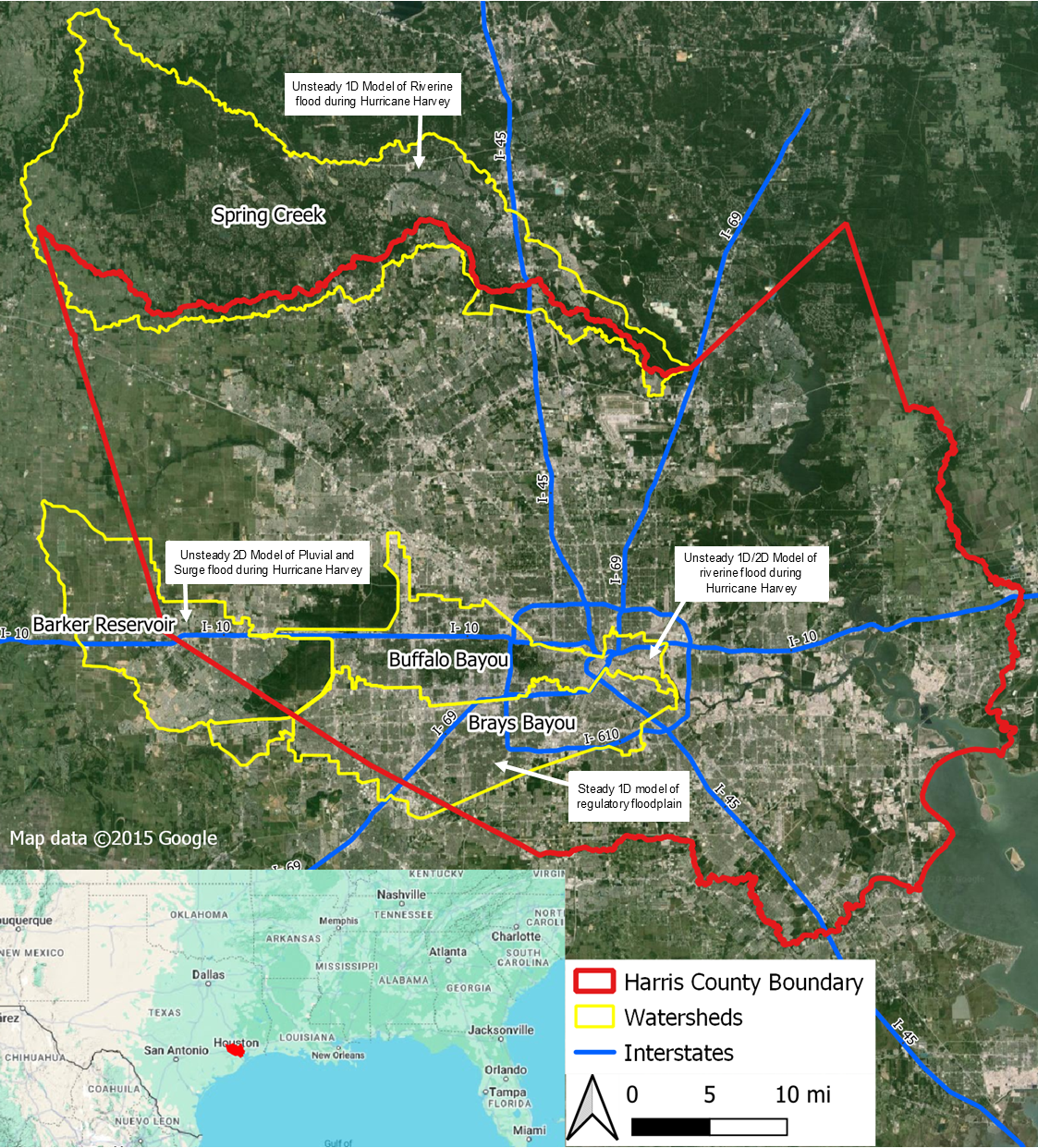}
	\caption{Relevant watersheds for hydrodynamic model setups in Harris County.}
	\label{harris}
\end{figure}

The developed DNN was applied to several regions of a 0.5-m 3D Elevation Program (3DEP) lidar survey \cite{lidarsurvey} of Houston, TX, USA. These measurements were then incorporated into 1D, 2D, and coupled 1D/2D hydrodynamic models to comprehensively test their performance across a variety of model domains and flood conditions. Houston, which is primarily located in Harris County, is a highly-urbanized low-relief environment on the Gulf of Mexico, and frequently experiences flooding. It's highly heterogeneous surface and rapid development make it an ideal location for this study. Figure \ref{harris} shows a satellite overview of Harris County overlaid with the relevant watersheds for this study's hydrodynamic models. Two flood events were used for this testing; one was the design flood event for regulatory floodplain mapping, and the other was Hurricane Harvey. The following subsections briefly describe Hurricane Harvey, and the model setups used to test the impacts of improved FFs data during pluvial, fluvial, and surge flooding. \par
\subsubsection{Hurricane Harvey}
Hurricane Harvey caused record-breaking amounts of rainfall in Houston between August 25th, 2017 and August 30th, 2017. Literature has found that hydrodynamic models of Hurricane Harvey are highly sensitive to \textit{Manning's n}, which makes it a crucial hydraulic parameter \cite{dullo2021}. Therefore, Hurricane Harvey is a challenging and useful benchmark to test the PC-based \textit{Manning's n} measurements. At the time of writing, Hurricane Harvey is the most significant flood event for both of Houston's major reservoirs (Addicks and Barker), causing uncontrolled water releases through emergency spillways, and subsequent controlled releases to decrease water elevation. During Harvey, the reservoir pools exceeded government property and flooded upstream neighborhoods, while the controlled releases contributed to downstream flooding \cite{HarveyReservoirs}. 
\subsubsection{1D Modeling}
\textit{Brays Bayou}\par
A steady 1D HEC-RAS model was developed for Brays Bayou (HCFCD ID D100-00-00) in Houston, TX. The model was set up using publicly-available cross section and design storm information from the Houston M3 platform maintained by the Harris County Flood Control District \cite{M3}. This is the same data used to generate FEMA flood hazard areas (i.e., 100-year floodplains), and contains both terrain and \textit{Manning's n} values for it's cross sections. These \textit{Manning's n} values are selected by engineering judgement using two sets of guidelines \cite{PCPM2019} \cite{H&HM2009}. \par
Using the trained DNN, \textit{Manning's n} measurements were obtained from a 3DEP lidar survey flown by AECOM in 2018 \cite{lidarsurvey}. To incorporate these spatial roughness measurements into 1D cross sections, Python scripts were developed to cluster spatial groupings of cross section stations using K-Means \cite{lloyd1982k-means}, and compound their Manning's roughnesses using the Horton-Einstein method \cite[equation 6-17]{chow1959}\cite{horton1933, einstein1934hydraulische}. This procedure was required both for numerical stability and due to HEC-RAS' limitations on the amount of horizontally-varied \textit{Manning's n} (which is 20 per cross section).  The Python scripts are provided at \href{https://github.com/f-haces/LidarManning}{https://github.com/f-haces/LidarManning}. \par
Land cover data was obtained from the National Land Cover Dataset (NLCD) \cite{NLCD}. HEC-RAS has capabilities to directly estimate \textit{Manning's n} values derived from land cover for cross sections, however, doing so removes any in-channel information. As such, two models were set up; one with the regulatory model in-channel \textit{Manning's n} and the mean values from NLCD in the overbank regions, and another with the channel \textit{Manning's n} entirely derived from land cover processed through the Horton-Einstein method. This resulted in four model setups: the original regulatory model, the one based on the lidar measurements, and the two for land cover. All 1D models in Brays Bayou were set to predict the channel flow corresponding to a 100-year flood event, which was already defined in the original model files \cite{M3}.\par
\textit{Spring Creek}\par
A series of unsteady 1D HEC-RAS models was developed for Spring Creek (HCFCD ID J100-00-00) to assess riverine flooding conditions. Spring Creek is a very different hydrodynamic environment to the other watersheds shown in Figure \ref{harris}, with a wetland-derived headwater floodplain and large undeveloped surfaces \cite{SpringCreek}. As such, it was chosen to provide further testing for the presented lidar measurements of \textit{Manning's n} during Hurricane Harvey. Custom cross sections were created for a portion of Spring Creek, with channel elevations being burned into the underlying terrain from cross sections obtained from the regulatory model \cite{M3}. \par
\begin{figure}[!htb]
	\centering
	\includegraphics[width=1\textwidth]{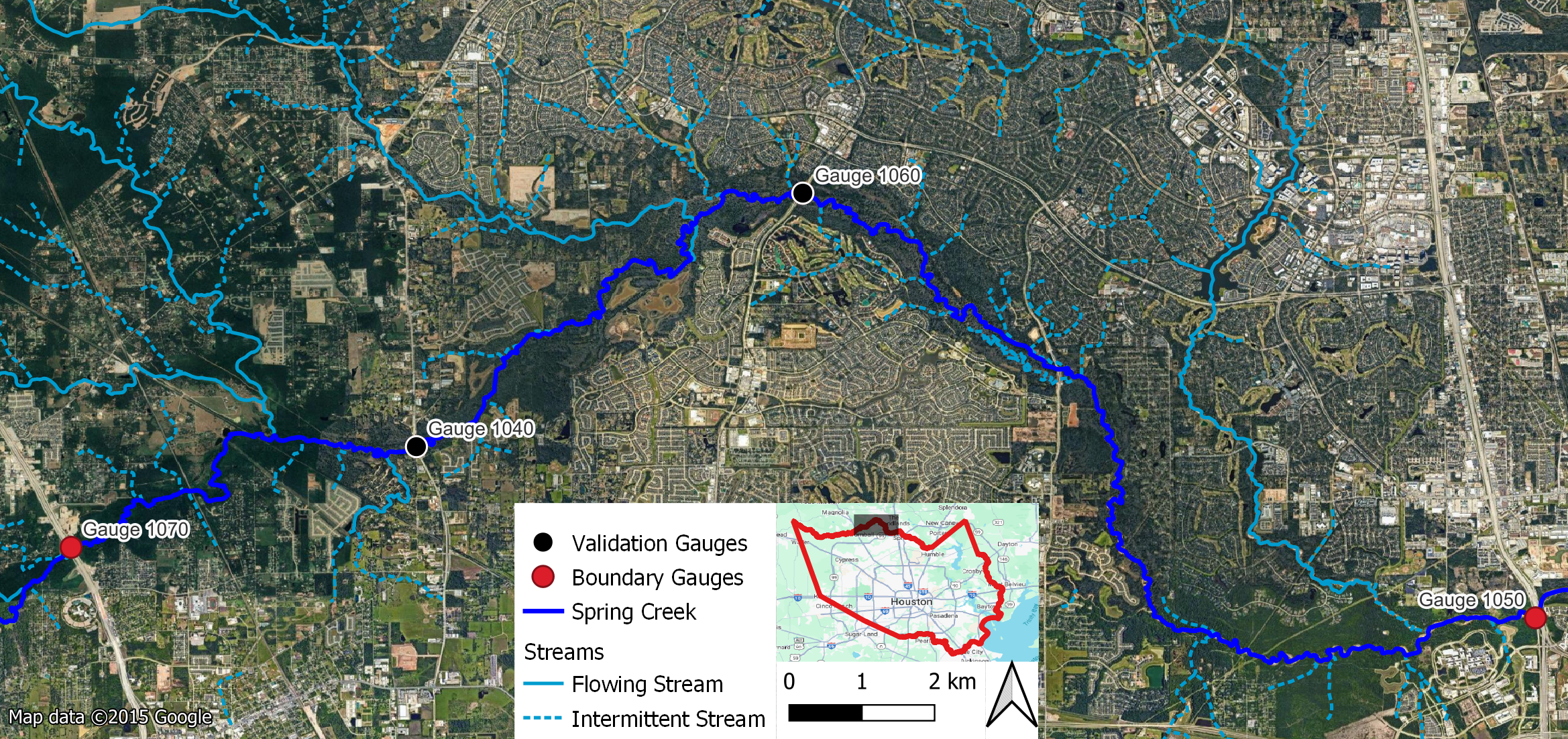}
	\caption{1D Model Setup to study Riverine Flooding in Spring Creek.}
	\label{springcreek}
\end{figure}

 
Two unsteady 1D hydrodynamic models were developed, one with the land cover values, and one with the lidar values from the 2018 3DEP lidar survey. This model setup is shown in Figure \ref{springcreek}, with four gauges used in this region (two as boundary conditions, and two as validation). All gauge data was obtained from Harris County Flood Warning System (HCFWS) between August 26th, 2017, and August 31st, 2017.  The streams flowing into Spring Creek are noted as influencing factors in this setup, however, since they were ungauged during Hurricane Harvey, they could not be included in the model.
\subsubsection{1D/2D Coupled Modeling}
\begin{figure}[!htb]
	\centering
	\includegraphics[width=1\textwidth]{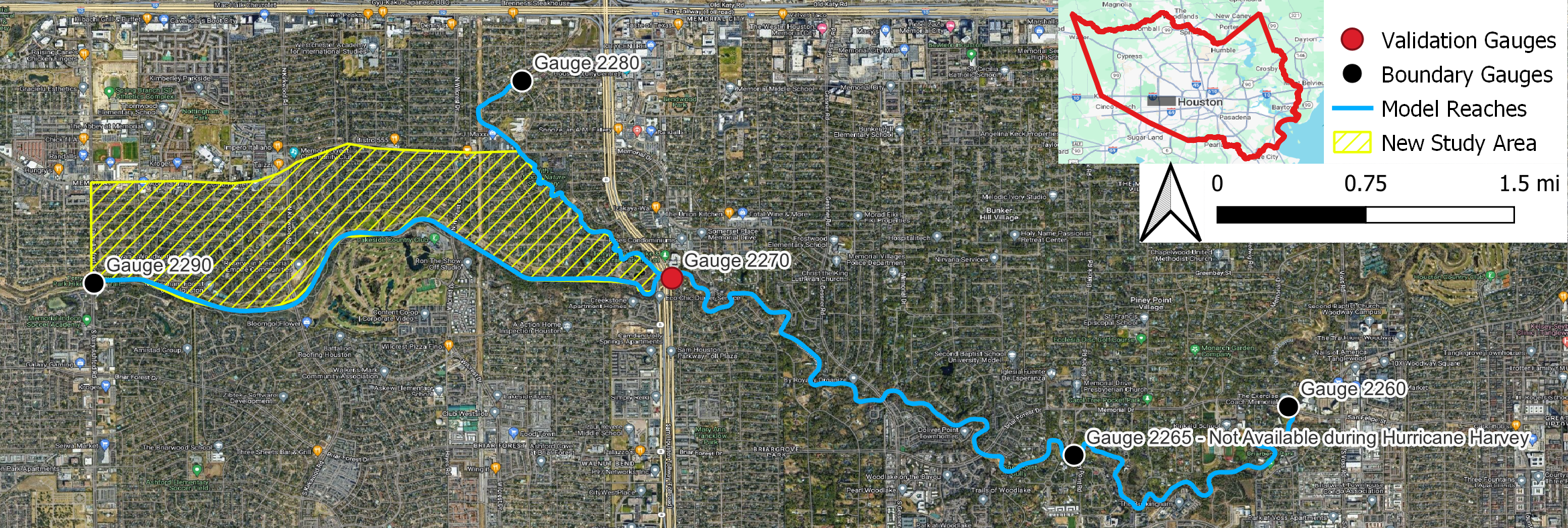}
	\caption{1D/2D Coupled Model Setup to study Riverine Flooding in Buffalo Bayou. }
	\label{BuffaloBayou}
\end{figure}

A coupled 1D/2D model was set up to further assess riverine flooding conditions during Hurricane Harvey in a portion of Buffalo Bayou (HCFCD ID W100-00-00), which was impacted by the controlled releases of both the Addicks and Barker Reservoirs. Channel flow was modeled by setting the upstream boundary condition as the flow gauge at the Dairy Ashford Road (Site 2290), and the downstream boundary condition the flow gauge at San Felipe Dr (Site 2260). An additional gauged stream (HCFCD ID W156-00-00) was added to the model, with a 2D flow area being defined to study flooding upstream of it's confluence with Buffalo Bayou. The domain allowed for one validation gauge (Site 2270), which is at the outflow of the 2D area. This setup is shown in figure \ref{BuffaloBayou}. Gauge data at irregular time intervals between August 26th, 2017, and August 31st, 2017, was obtained from HCFWS. The model's terrain was obtained from the 2018 3DEP lidar survey, and channel bathymetry from regulatory models was burned in. Coupling between the 1D and 2D regions was achieved by a lateral weir (as is standard practice in HEC-RAS \cite{coupledguide}). \par

Two model setups were created, one with land cover data, and another with lidar measurements from the 2018 3DEP survey. The mean of the left and right sides of the channel were taken for \textit{Manning's n} values. Using the Horton-Einstein method to compound \textit{Manning's n} was deemed unnecessary due to the short cross section lengths, which were required to couple 1D and 2D flows.  In the 2D areas, the \textit{Manning's n} values were taken directly from the underlying rasters in both models. Validation data was sought from FEMA National Flood Insurance Program data, aerial imagery during Hurricane Harvey, and Harris County Flood Control District.
\subsubsection{2D Modeling}
\begin{figure}[!htb]
	\centering
	\includegraphics[width=1\textwidth]{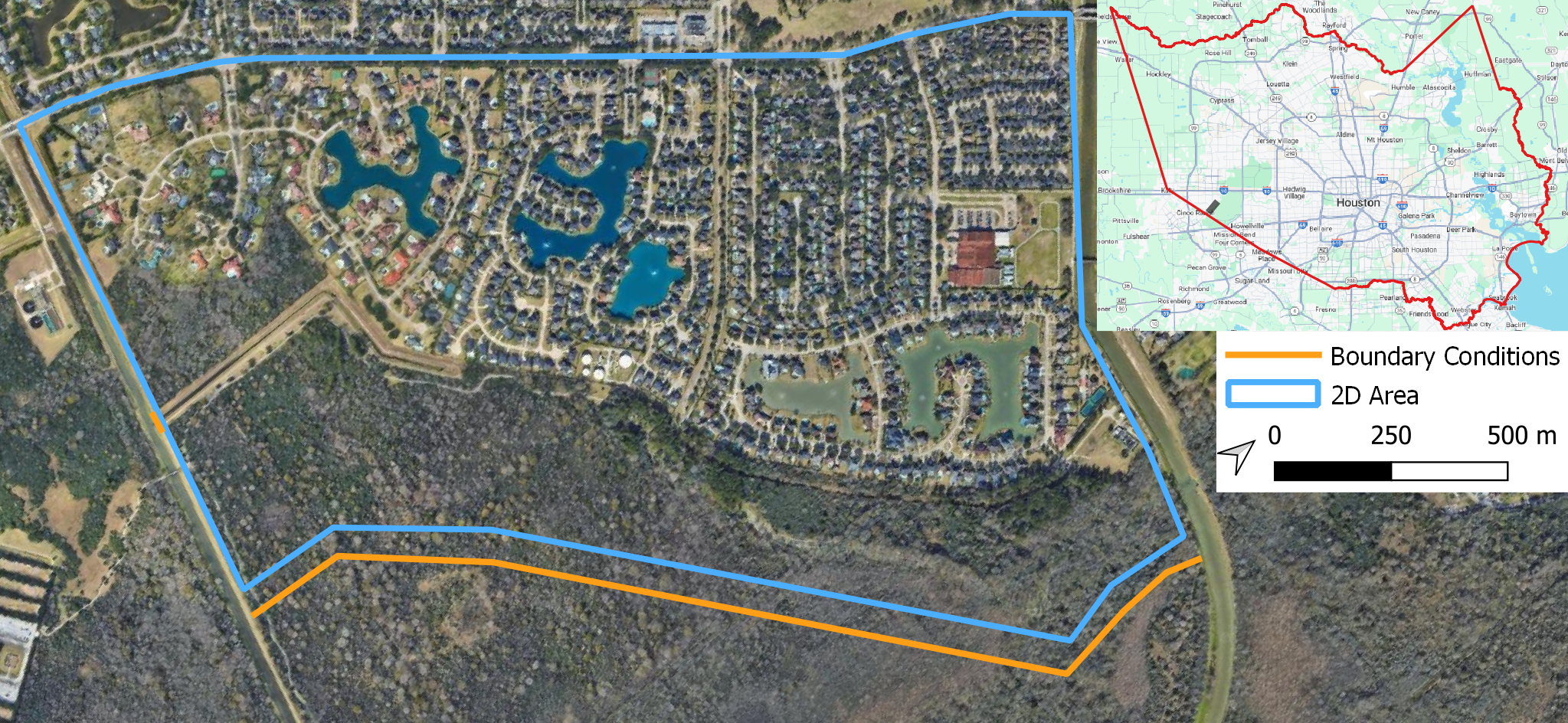}
	\caption{2D Model Setup to study surge and pluvial flooding upstream of Barker Reservoir. }
	\label{Buckingham}
\end{figure}

A 2D model was created to assess surge and pluvial flooding upstream of Barker Reservoir, in an area that was impacted by rising pool levels. Terrain data was collected from the 2018 3DEP lidar survey \cite{lidarsurvey}, and used to generate \textit{Manning's n} measurements. Rising pool level data was obtained from HCFWS' gauge at Barker Dam (Site 2010), while rainfall data was collected at Site 2020 (which was closer to the model domain). This model setup is shown in Figure \ref{Buckingham}. \par
\section{Results}
\begin{figure}[!htb]
	\centering
	\includegraphics[width=1\textwidth]{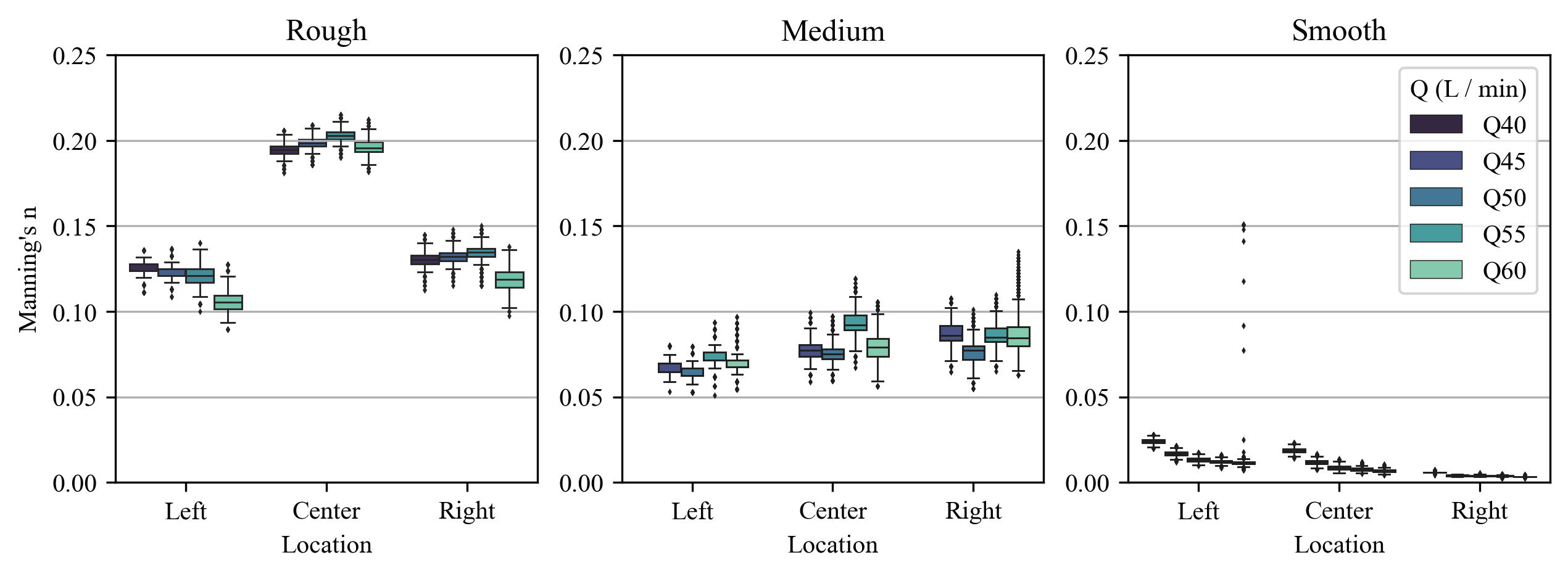}
	\caption{\textit{Manning's n} measurements across different surfaces. The boxplots represent variation in instantaneous water elevation measurements across the measurement time. }
	\label{n_measurement}
\end{figure}

The laboratory measurement of \textit{Manning's n} provided some interesting insights into the surfaces. The measurements obtained for each concrete slab are shown in Figure \ref{n_measurement}. For the smoothest surface, there was a slight trend in the Manning's n measurement dependent on the flow depth. This has been reported in the literature before (e.g., \cite{Garrote2021}). The mean values for all experiments were used for each measurement region. There was also significant lateral variation in the \textit{Manning's n} value for the roughest surface. This was due to observed water flow non-uniformity during experimentation. As such, the value for the center measurement zone was used for both the left and right regions during data augmentation.\par
\begin{figure}[!htb]
	\centering
	\includegraphics[width=1\textwidth]{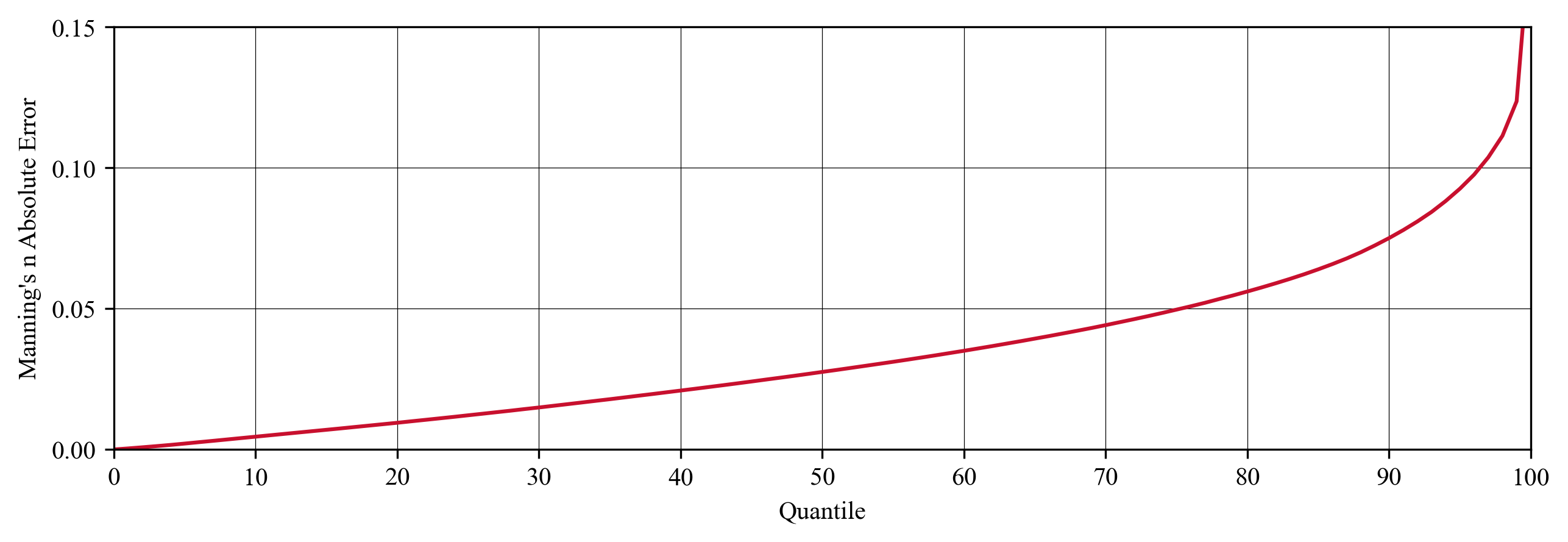}
	\caption{DNN performance for the test dataset (n=804,000). }
	\label{fit}
\end{figure}

After training, the DNN's performance was assessed using samples from the test dataset, which is shown in Figure \ref{fit}. The mean and median absolute errors were 0.035 and 0.028, respectively. The performance of the trained DNN was found to be highly sensitive to point cloud scale; that is, processing spatial extents significantly different than those used during laboratory training resulted in a rapid degradation in performance. As such, to generate \textit{Manning's n} measurements, all point clouds were grouped spatially in rectangular radii of 1-m, on which further processing was then performed.\par
\begin{figure}[!htb]
	\centering
	\includegraphics[width=1\textwidth]{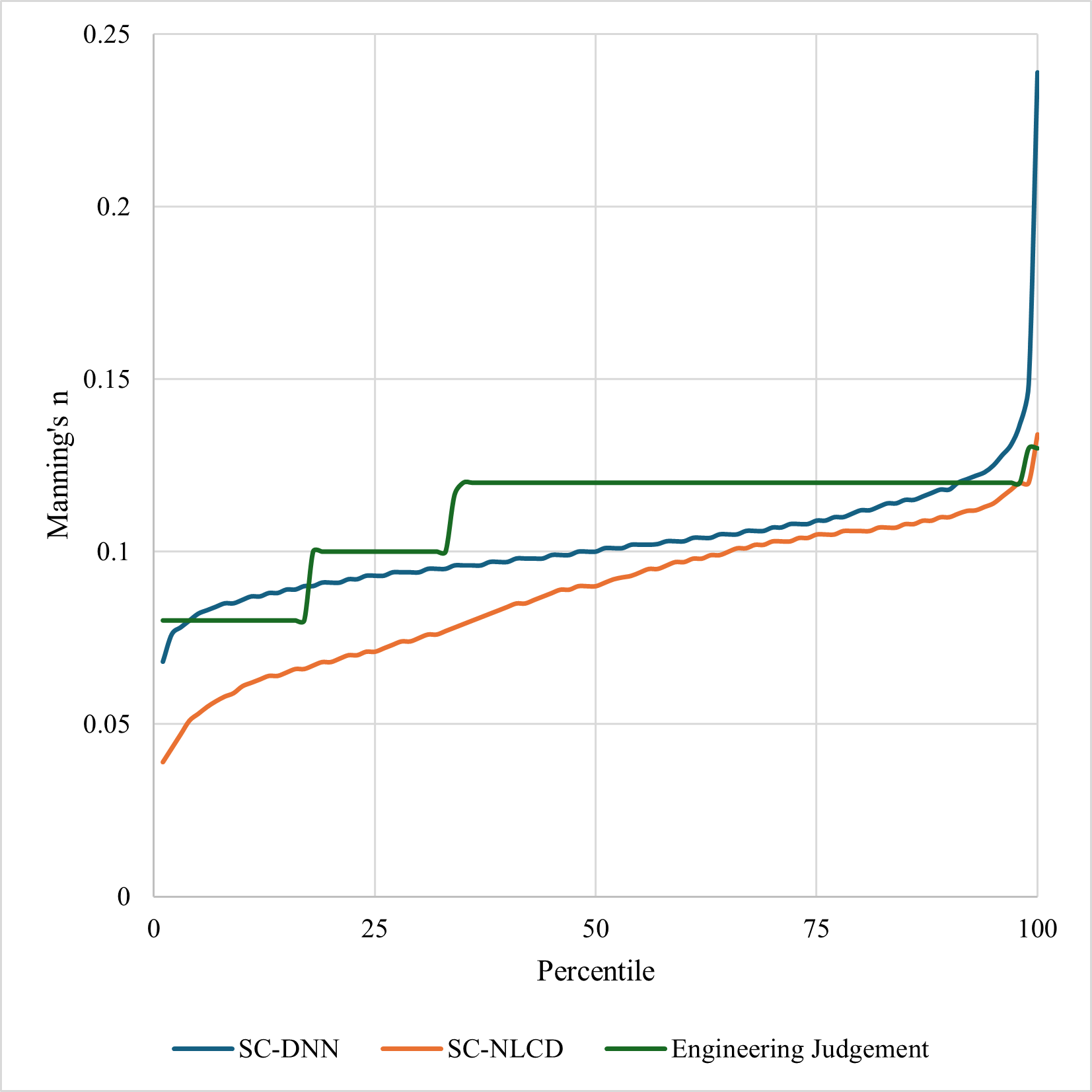}
	\caption{Comparison of \textit{Manning's n} values obtained from DNN, NLCD, and Engineering Judgement. Engineering judgement values are solely from overbanks regions. }
	\label{manningcomp}
\end{figure}

A comparison between the values obtained from the lidar PCs, land cover, and Engineering Judgement for the steady 1D regulatory model of Brays Bayou is shown in Figure \ref{manningcomp}. For this figure, Engineering Judgement values were taken from the overbanks regions, as neither the NLCD or lidar PC contain in-channel data. Taking the mean \textit{Manning's n} of each cross section, the lidar measurements showed a 37.6\% improvement in Root Mean Square Error over the NLCD ($RMSE_{DNN}=0.0187$, $RMSE_{NLCD}=0.0299$, $n=387$). Note that the stair-stepping in Figure \ref{manningcomp} indicates that Engineering Judgement values for the overbank regions were selected in intervals of $n=0.02$.\par
\begin{figure}[!htb]
	\centering
	\includegraphics[width=1\textwidth]{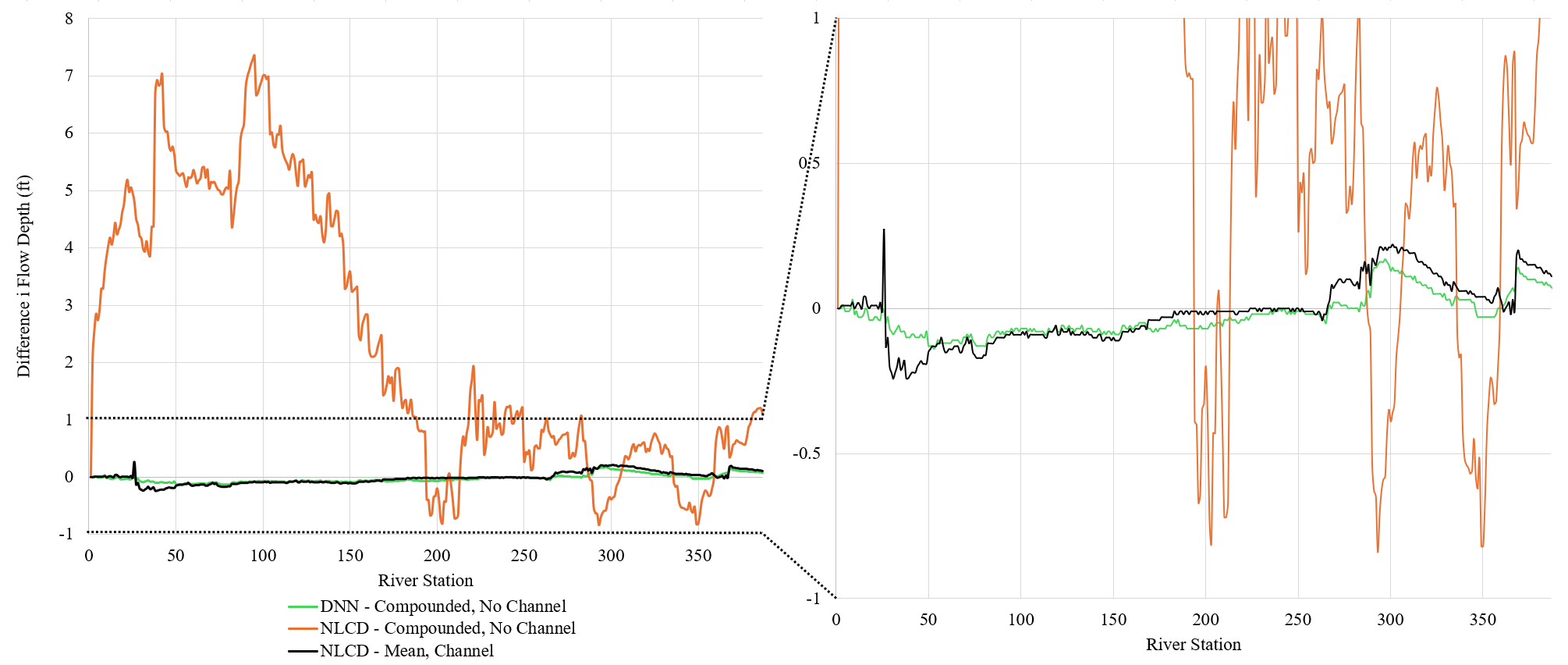}
	\caption{Difference in flow depth across various friction conditions for the 1D regulatory model. The reference flooding depth is that calculated using the original model \textit{Manning's n} from HCFCD. \cite{M3}}
	\label{1D}
\end{figure}

Figure \ref{1D} shows the changes caused by different FFs to the regulatory 100-year 1D HEC-RAS model of Brays Bayou's. Using \textit{Manning's n} values purely derived from land cover caused an exaggerated increase in flow depth at some points of the bayou. While including channel information and using the mean value somewhat improved on these issues, important differences and flow shocks are still present. The statistics shown in Table \ref{tab:1D} reveal 24.9\% and 37.0\% improvements in mean and maximum water depth deviations, respectively, by using the DNN's measurements over land cover when compared to the Engineering Judgement values. Note that in-channel values were not artificially added to the PC measurements, however, the lidar PC has no points in the channel, and no-data values for \textit{Manning's n} were set to the lowest value in the experimental measurement range (0.025) prior to compounding.\par
\begin{figure}[!htb]
	\centering
	\includegraphics[width=1\textwidth]{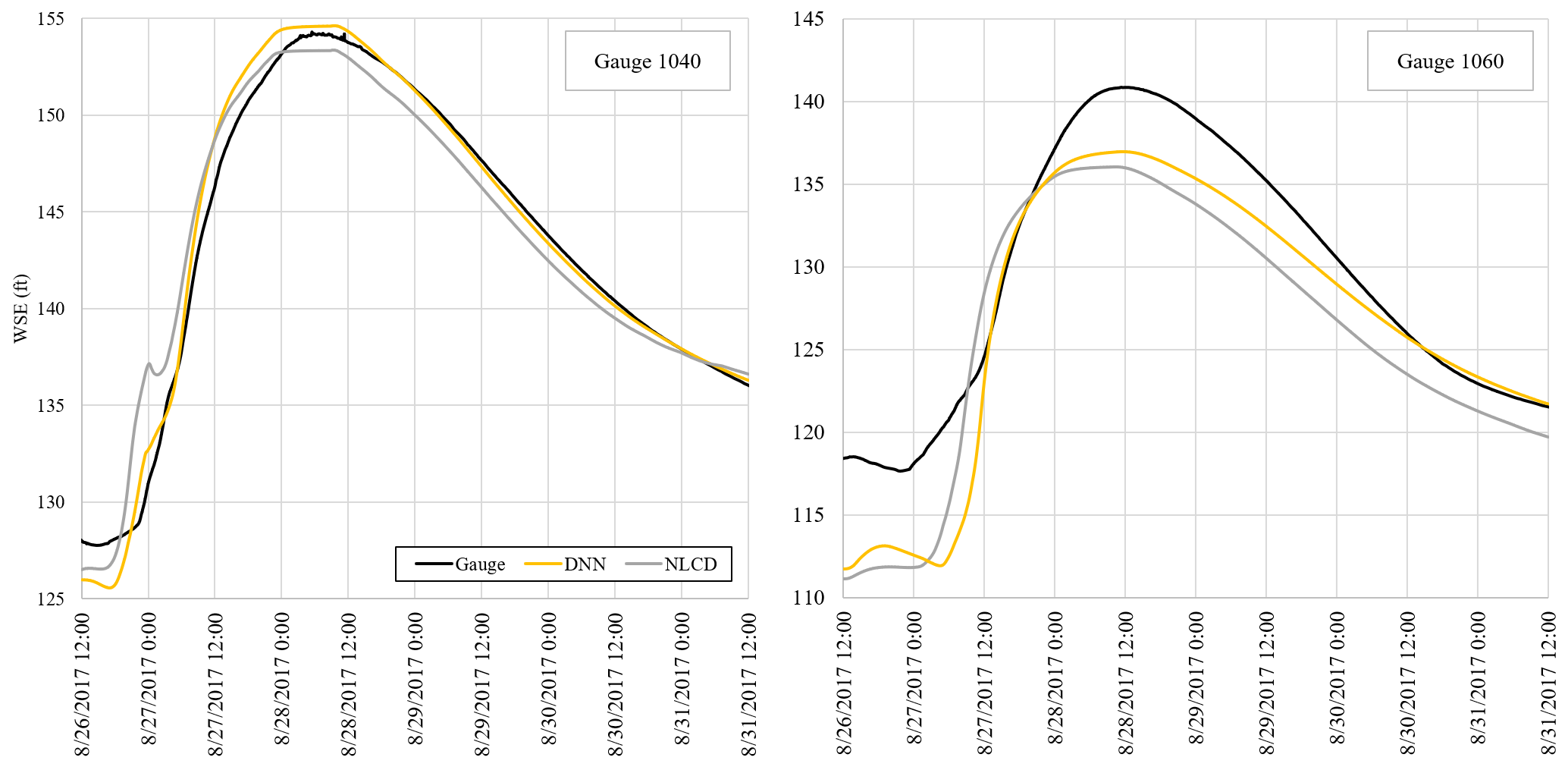}
	\caption{Comparison of 1D unsteady model timeseries for Spring Creek at validation gauges.}
	\label{SpringGauges}
\end{figure}

\begin{table}[!htb]
\caption{\label{tab:1D}Deviation in water depth as compared to original  1D regulatory HEC-RAS model.}
\begin{center}
\begin{tabular}{l|l|l|l|}
\cline{2-4}
                                        & Mean (ft) & 99th (ft) & Max (ft)  \\ \hline
\multicolumn{1}{|l|}{NLCD (Compounded, No Channel)} & 2.48& 7.11& 7.36\\ \hline
\multicolumn{1}{|l|}{NLCD (Mean, Channel)}    & 0.08& 0.21& 0.27\\ \hline
\multicolumn{1}{|l|}{DNN - (Compounded, No Channel)}               & 0.06& 0.16 & 0.17 \\ \hline
\end{tabular}
\end{center}
\end{table}
The timeseries obtained from validation gauges for the unsteady 1D models of Spring Creek during Hurricane Harvey are shown in Figure \ref{SpringGauges}. As illustrated in Figure \ref{springcreek}, there are several ungauged streams that flow into Spring Creek, complicating hydraulic assessment. Their impact on the models vary across locations; models show good agreement with real-world observations of Gauge 1040, while having markedly lower peaks but similar drawdowns at Gauge 1060. Importantly, the lidar measurements result in better fit than the NLCD values for the Water Surface Elevation (WSE) timeseries at both gauges; these metrics are summarized in Table \ref{tab:SpringFit}.

\begin{figure}[!htb]
	\centering
	\includegraphics[width=1\textwidth]{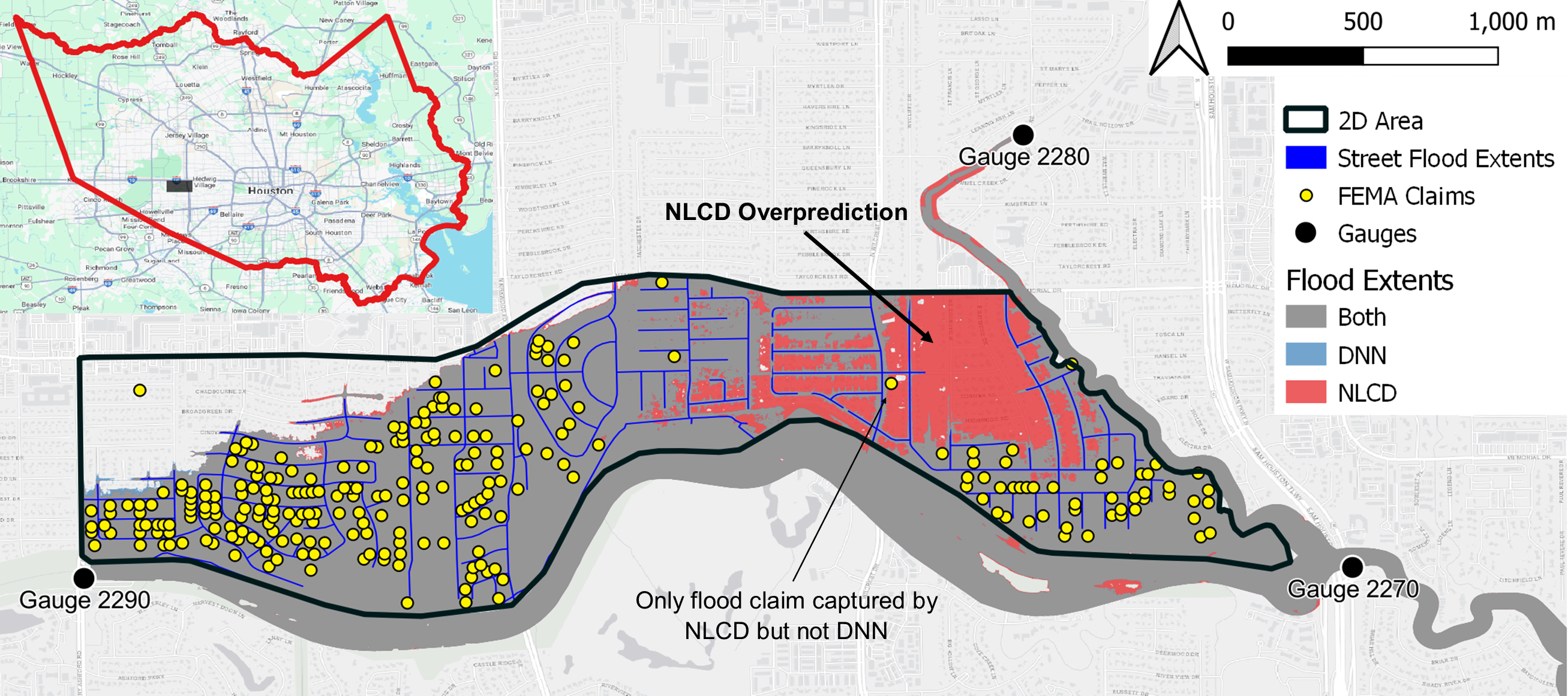}
	\caption{Comparison of flood extents at time of imagery (08/30/17 10:00 AM) in 1D/2D coupled models of Buffalo Bayou during Hurricane Harvey. FEMA National Flood Insurance Program claims in 2D area are also shown.} 
	\label{hybrid}
\end{figure}

\begin{table}[!htb]
\centering
\caption{Fit metrics at validation gauges for 1D unsteady model of Spring Creek. The DNN predicts }

\begin{tabular}{l|c|c|c|}
\cline{2-4}
                                  & \begin{tabular}[c]{@{}c@{}}Nash Suttcliffe\\ Efficiency\\ (Higher is better)\end{tabular} & \begin{tabular}[c]{@{}c@{}}Root Mean \\ Square Error\\ (Lower is better)\end{tabular} & \begin{tabular}[c]{@{}c@{}}Final Absolute \\ Difference\\ (Lower is better)\end{tabular} \\ \hline
\multicolumn{1}{|l|}{Gauge 1040 - Lidar}  & 0.983                                                                                   & 1.038                                                                              & 0.255                                                                        \\ \hline
\multicolumn{1}{|l|}{Gauge 1040 - NLCD} & 0.950                                                                                   & 1.765                                                                              & 0.561                                                                        \\ \hline
\multicolumn{1}{|l|}{Gauge 1060 - Lidar}  & 0.800                                                                                   & 3.479                                                                              & 0.175                                                                        \\ \hline
\multicolumn{1}{|l|}{Gauge 1060 - NLCD} & 0.713                                                                                   & 4.171                                                                              & 1.825                                                                       \\ \hline
\end{tabular}

% 1 - 1.038 /1.765 = 41.2
% 1 - 3.479 / 4.171 = 16.6
% 1 - 0.255 / 0.561 = 54.5
% 1- 0.175 / 1.825 = 90.4

\label{tab:SpringFit}
\end{table}
A comparison of the 1D/2D coupled models' flood extents is shown in Figure \ref{hybrid}. Occlusions in the contemporaneous NOAA imagery \cite{NOAAHarvey} prevent the accurate delineation of flood extents, however, the street network remains largely visible. As such, flood extents for the street network in the imagery were delineated and included as validation data. FEMA National Flood Insurance Program Claims during Hurricane Harvey in the study area are also included. Several interesting observations can be made from the modeled flood extents. The lidar \textit{Manning's n} measurements generally result in less extensive flooding. Moreover, NOAA imagery shows better agreement with that model's flooding than with NLCD values, particularly in the eastern portion of the study area (as shown by the overprediction in Figure \ref{hybrid}). This is corroborated by fit metrics inside the street network, with the NLCD having lower Intersection over Union (0.867 and 0.921, respectively) and F1 scores (0.94 and 0.97, respectively). Out of a total of 253 paid out flood claims, lidar and NLCD capture 251 and 252 respectively, covering 99.8\% and 99.9\% of paid out damages for the 2D area. \par
%Intersection over Union (IoU): 0.9207820508578558 Intersection over Union (IoU): 0.8674990517078065 
% F1: 0.9672111927534132 F1: 0.9399162404938

\begin{figure}[!htb]
	\centering
	\includegraphics[width=5in]{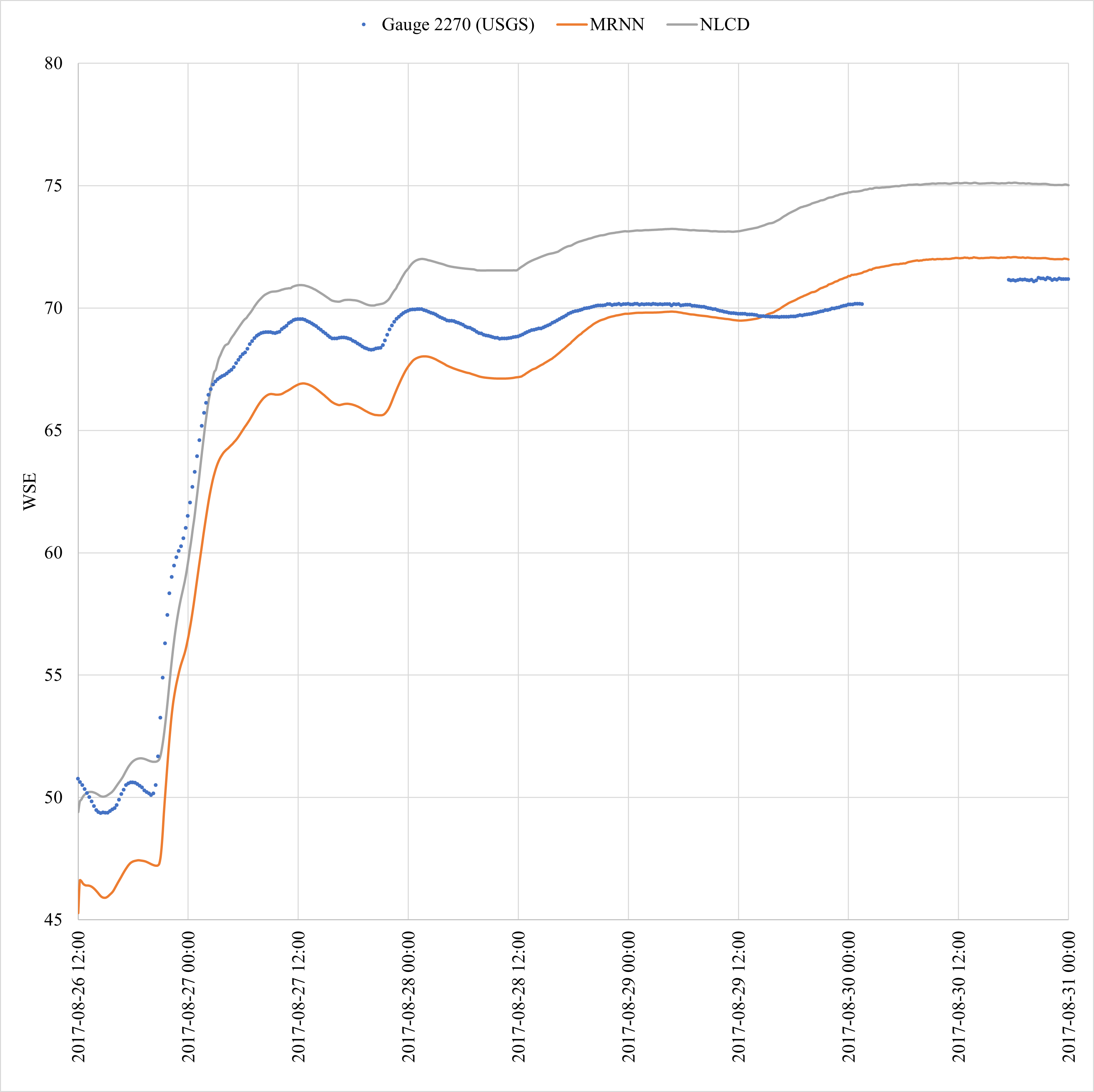}
	\caption{Flood elevation difference in 2D model of subdivisions upstream of Barker Reservoir using the HEC RAS Full Momentum Equations. While surge flooding shows no change in extents or depth, street networks further away from reservoir show clear differences in rainfall flooding depth.}
	\label{BuffaloGauge}
\end{figure}

Figure \ref{BuffaloGauge} compares the WSE at Gauge 2270 (which was used as a validation gauge) with the modeled WSEs for the lidar and NLCD's values. The NLCD's timeseries reveals deeper WSEs at the gauge location than both the real-world observations and the lidar measurements, corroborating the more extensive flooding predicted in the 2D area. The lidar measurements result in a closer match to the gauge observations than the NLCD's values, with a higher Nash-Suttclife Efficiency (0.81 and 0.60, respectively) and lower Root Mean Square Error (2.41 and 3.52, respectively). The difference to the gauge at the final modeled timestep (which, in this model, is relevant to assess maximum flood extents due to backwater flooding) was also higher for NLCD than lidar (1.7 ft and 4.7 ft, respectively).

\begin{figure}[!htb]
	\centering
	\includegraphics[width=1\textwidth]{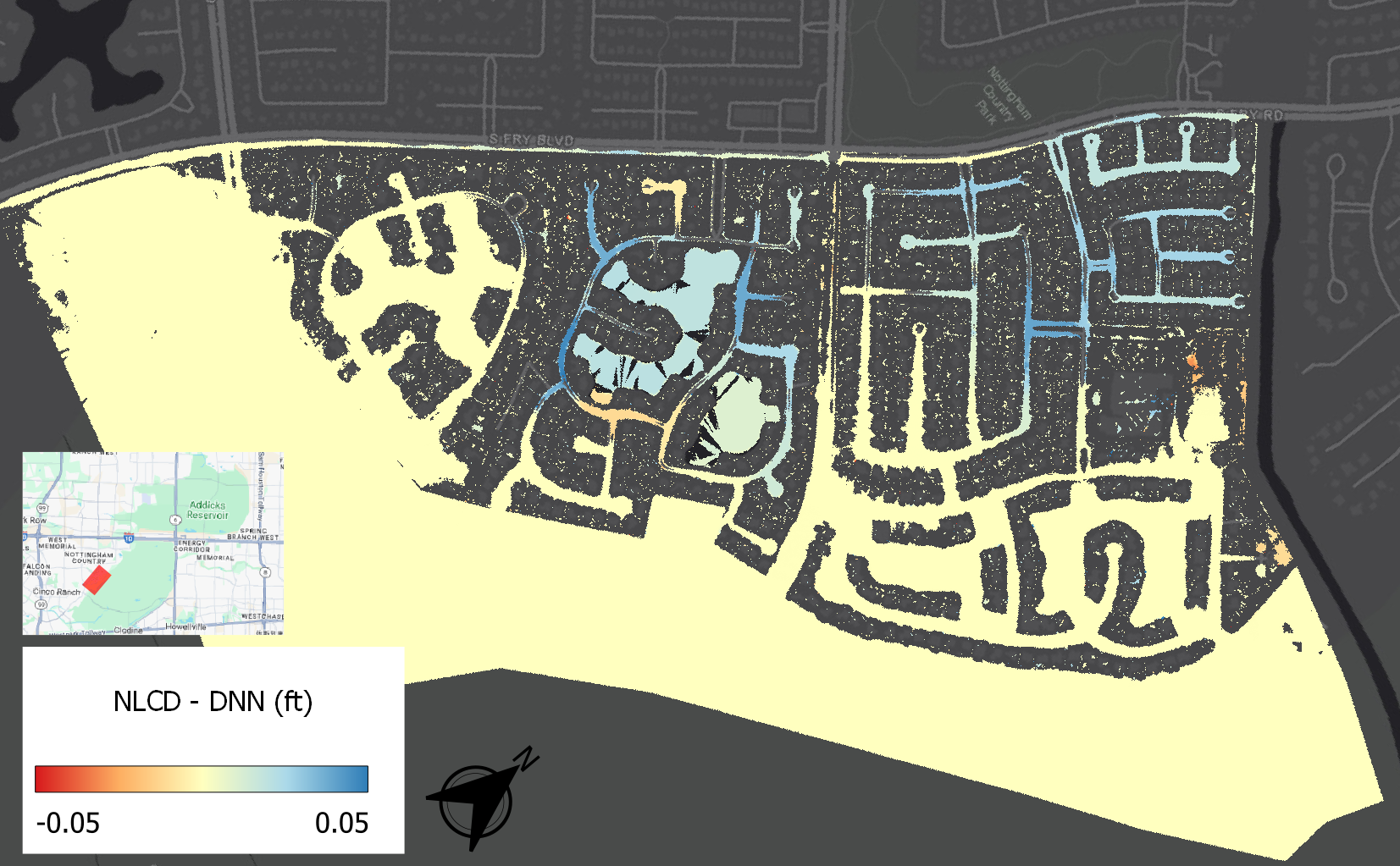}
	\caption{Flood elevation difference in 2D model of subdivisions upstream of Barker Reservoir using the HEC RAS Full Momentum Equations. While surge flooding shows no change in extents or depth, street networks further away from reservoir show clear differences in rainfall flooding depth.}
	\label{2D}
\end{figure}

When comparing the 2D models upstream of the Barker reservoir, striking similarities in surge-derived flooding are apparent between the NLCD and lidar \textit{Manning's n} values. In fact, water depth at all locations flooded by the rising reservoir is exactly the same, regardless of which HEC-RAS equation set was used (both the Full Momentum Equations and Diffusion Equations were tested). Flooding extents are also similar for street-level flooding derived from rainfall, however, flooding depths change across locations. When using the NLCD's \textit{Manning's n} estimates, total flooding volumes were 0.52 and 0.35 acre-ft greater for the Full Momentum Equations and the Diffusion Equations, respectively. The flooding differences across the two surface roughness datasets are shown in Figure \ref{2D}.
\section{Discussion}
The lidar PC provides a closer match to engineering judgment than \textit{Manning's n} values derived from land cover, improving the mean \textit{Manning's n} for each cross section by up to 37.6\%, while allowing measurements at higher resolution. This suggests that the DNN can be used to better standardize the measurement of FFs across locations, removing inherent biases and uncertainties of judgement-based values while also providing a low-cost method for widespread implementation. The PC measurements provide a better match to steady regulatory models than values derived from land cover. They also simplify model setup by reducing manual inputs; the use of land cover required significant modifications to in-channel roughness and changes to \textit{Manning's n} compounding methods to remove exaggerated increases in flow depths. The lidar measurements, however, were simply postprocessed with the Python scripts (available at \href{https://github.com/f-haces/LidarManning}{https://github.com/f-haces/LidarManning}) and provided better agreement across all models. \par

The presented models generally improve on previously-available flooding estimates during Hurricane Harvey. To the knowledge of the authors, this is the first manuscript to develop direct comparisons between  hydrodynamic models and contemporaneous airborne imagery for Hurricane Harvey. In the unsteady 1D model of Spring Creek, the use of PC measurements resulted in better fit with the validation gauges. In the 1D/2D coupled model, they also provided a closer match to the validation gauge, while simultaneously improving fit to street flood extents derived from airborne imagery. The PC values also result in a decrease of predicted flood extents in the 1D/2D coupled model, improving flood prediction in the 2D area while only decreasing the coverage of flood insurance claims by one (representing 0.01\% of damages in the 2D area). The inclusion of validation gauges as boundary conditions would improve the performance of the models based on either source for \textit{Manning's n} values. However, doing so would remove critical sources for validation in this study. \par

FFs were found to significantly impact 1D and 2D model domains under riverine and pluvial flooding, while surge flooding was unaffected. This is perhaps unsurprising; bottom friction is most significant in gravity-dominated flow. However, the results from Spring Creek and Buffalo Bayou suggest an interesting caveat to this phenomenon; the severity of FF's impacts on riverine flood models may also be dependent on downstream conditions. Backwater flooding changes the composition of gravity-dominated flow, behaving similarly to surge flooding. This is shown by the relative performances of both sets of \textit{Manning's n} in Spring Creek and Buffalo Bayou. The former did not experience backwater flooding and had lower final WSE differences (0.255ft and 0.561ft for Gauge 1040) despite being affected by ungauged streams. The latter did experience backwater flooding and had much higher WSE differences (1.7ft and 4.7ft for Gauge 2270). While this is an imperfect comparison due to the change in flood environments and model setups, such a drastic difference in model performance across flow conditions warrants further study. \par

This research extends the literature findings on the significance of FFs to hydrodynamic models, studying their impact to different kinds of flood events. Moreover, this research contributes a new, readily-applicable avenue to obtaining high-resolution FF values that have been shown to improve 1D, 2D, and coupled models. The developed DNN can be applied to other PCs using the code available at \href{https://github.com/f-haces/LidarManning}{https://github.com/f-haces/LidarManning}. At the time of writing, USGS's 3DEP covers most of the conterminous United States with high-resolution lidar data, making it possible to obtain high-resolution \textit{Manning's n} values. It is noted that most high-resolution hydrodynamic flood modeling already relies on such data for terrain inputs. PCs are collected with a variety of different sensors, scanning patterns, quality levels, and horizontal spacings, which can introduce heterogeneity into the DNN's measurements. As such, further testing of the capabilities of the presented DNN is needed. Nonetheless, this research demonstrates the great potential of DNNs to measure FFs from PC data in a distributed scale, with wide-ranging applicability.\par
% \subsection{Broader Implications for Flood Modeling}
\subsection{Future Research Directions}
The validation of hydrodynamic models is challenging. The model domains for this study were selected by balancing the significance of \textit{Manning's n} and the availability of data to use as validation. Many data sources were found to be incomparable to this study's models. While alternative modeling results for Hurricane Harvey are available in the literature (notably, \cite{wing2019, dullo2021, garcia2020, sebastian2021}), they generally do not encompass the same study areas, or are of much coarser resolution, whereas alternative sources of ancillary data were found to be inaccurate. For these reasons, it is challenging to validate the important changes to pluvial flooding volumes observed in this study. The difficulty is two-fold: HEC-RAS has no stormwater infrastructure modeling capabilities (as such, imagery is not applicable for validation of pluvial flooding extents), and the overall volume change is caused by small compounded changes in flooding depth, which are difficult to quantify. Nonetheless, such changes are a crucial consideration for stormwater conveyance infrastructure, even playing a role during Hurricane Harvey in the 2D model's study area; imagery shows that some stormwater inlets were overwhelmed by flooding, while other were not. \par
The presented DNN can be used on PCs to provide better FFs at higher resolutions than land cover. However, there is a wider ongoing discussion in the literature regarding the use of FFs in flood models. There are important well-documented limitations to using FFs to estimate friction in flood modeling. For example, FFs were originally designed for channels and pipes \cite{manning1890}, not overland flow. Moreover, FFs are frequently calibrated during modeling, making them prone to serve as lumped variables that account for external flow factors \cite{Lane2005, Whatmore2011}. Pragmatically, more complex friction formulations require both a reliable measurement technique, along with widespread implementation in modeling platforms. The presented DNN addresses the former; while in this study it was made to measure \textit{Manning's n}, it would be relatively straightforward to alternatively or in conjunction measure several friction-related parameters in a more complex friction-momentum formulation for flood models.\par 
This manuscript focused on illustrating the potential improvements enabled by PCs to friction estimation in flood modeling, thereby offering the opportunity for numerous avenues in which to expand this research. For example. further laboratory testing (particularly at different slopes) would provide more training data, which could improve DNN measurement performance. Testing under different flood events or other locations would provide insight into performance gaps as well. Moreover, while DNN architecture and hyperparameter tuning were performed in this study, it is possible that there are optimizations available within the presented architecture, or improved performance with alternative architectures. Further advancements can also be made based on this DNN; for example, it could serve as a base for a new architecture with transformation networks to prevent the observed scale dependence during inference. Finally, while this study showed the effectiveness of using multi-sensor PCs for training a single DNN, it is likely that improved data augmentation from generative deep learning would provide even better performance with a comparable amount of data. 

\section{Conclusion}
Hydrodynamic modeling relies on friction factors (FFs) to calculate momentum losses and simulate flood conditions. However, reliable measurements for FFs are difficult to obtain, and flood models often use surrogate remotely-sensed data with known drawbacks. This research presented a new method to measure \textit{Manning's n} from Point Clouds (PCs) using a DNN trained on data obtained from laboratory experiments. The implementation of the resulting \textit{Manning's n} values in hydrodynamic models showed improvements in both 1D and 2D domains. PC-based measurements were shown to more closely match engineering judgement than values based on land cover, providing a standardized and tested alternative for more accurate FF measurements. Fluvial flood models were shown to more closely match airborne imagery and gauge data for Hurricane Harvey when using the PC-based \textit{Manning's n} measurements. Pluvial models, while more challenging to validate, showed changes in the volume of street flooding, which warrants further study for its associated implications to stormwater infrastructure. Surge flooding was largely unaffected by changing FFs, however, potential linkages between surge physics and fluvial flooding were found. The presented DNN is made publicly available at \href{https://github.com/f-haces/LidarManning}{https://github.com/f-haces/LidarManning}, and can be readily applied to other study areas with lidar point clouds for further validation and testing. As such, this research presents a new method to apply extensively available PC data to measure FFs for hydrodynamics, avoiding sources for uncertainty in flood modeling and allowing more accurate predictions of storm events. 

\section{Author Contributions}
\textbf{Francisco Haces-Garcia}: Conceptualization, Methodology, Software, Validation, Formal Analysis, Investigation, Resources, Data Curation, Writing - Original Draft, Visualization\par
\textbf{Vasileios Kotzamanis}: Conceptualization, Methodology, Investigation, Resources, Writing - Review and Editing\par
\textbf{Craig Glennie}: Conceptualization, Methodology, Resources,  Writing - Original Draft, Supervision, Project Administration, Funding Acquisition\par
\textbf{Hanadi Rifai}: Conceptualization, Methodology, Resources, Writing - Review and Editing, Supervision, Project Administration, Funding Acquisition\par
\section{Acknowledgements}
The authors extend their sincerest gratitude to the lab supervisor of South Park Annex, Scott Smart. Dimitrios Kalliontzis is also gratefully acknowledged. Dawson Glennie is also thanked for his expert craftsmanship. The support of the Department of Civil and Environmental Engineering at the University of Houston is also gratefully acknowledged. Partial funding for the research was provided by the National Science Foundation through a grant to the National Center for Airborne Laser and Mapping (NCALM, NSF Grant \#1830734), and funding from the Hurricane Resilience Research Institute at the University of Houston. 
\section{Declaration of AI and AI-assisted technologies in the writing process}
During the preparation of this work the authors did not use any AI technologies in the writing process.
\printbibliography[title={Bibliography}]

\end{document}